\documentclass{article}

\PassOptionsToPackage{numbers, compress}{natbib}

\usepackage[final]{neurips_2024}

\usepackage{amsmath} 
\usepackage[utf8]{inputenc} 
\usepackage[T1]{fontenc}    
\usepackage{url}            
\usepackage{booktabs}       
\usepackage{amsfonts}       
\usepackage{nicefrac}       
\usepackage{microtype}      
\usepackage[numbers]{natbib}
\usepackage[dvipsnames,table,xcdraw]{xcolor}
\definecolor{cvprblue}{rgb}{0.21,0.49,0.74}
\usepackage{graphicx}
\usepackage{caption}
\usepackage{arydshln}
\usepackage{tikz}
\usepackage{wrapfig}
\usepackage{float}
\usepackage{multirow}
\usepackage{pifont}
\usepackage{makecell}
\usepackage{subfigure}
\usepackage[pagebackref,breaklinks,colorlinks,citecolor=cvprblue]{hyperref}

\newcommand{\tablestyle}[2]{\setlength{\tabcolsep}{#1}\renewcommand{\arraystretch}{#2}\centering\small}

\title{R1-Omni: Explainable Omni-Multimodal Emotion Recognition with Reinforcement Learning}

\author{
\textnormal{Jiaxing Zhao}\quad 
\textnormal{Xihan Wei}\quad 
\textnormal{Liefeng Bo}\quad\\
{\normalsize{Tongyi Lab, Alibaba Group}}  \\
{\tt{\small{zjx244036@alibaba-inc.com}}}\\
\textbf{\normalsize{\url{https://github.com/HumanMLLM/R1-Omni}}}\vspace{-3mm}}

\begin{document}

\maketitle


\begin{abstract}
In this work, we present the first application of Reinforcement Learning with Verifiable Reward (RLVR) to an Omni-multimodal large language model in the context of emotion recognition, a task where both visual and audio modalities play crucial roles. We leverage RLVR to optimize the Omni model, significantly enhancing its performance in three key aspects: reasoning capability, emotion recognition accuracy, and generalization ability.
The introduction of RLVR not only improves the model's overall performance on in-distribution data but also demonstrates superior robustness when evaluated on out-of-distribution datasets. More importantly, the improved reasoning capability enables clear analysis of the contributions of different modalities, particularly visual and audio information, in the emotion recognition process. 
This provides valuable insights into the optimization of multimodal large language models.
\end{abstract}


    

\section{Introduction}
With the advent of DeepSeek R1~\cite{deepseekai2025deepseekr1incentivizingreasoningcapability}, the potential of Reinforcement Learning (RL) has garnered increasing attention from researchers working on large models. A key innovation introduced by DeepSeek R1 is Reinforcement Learning with Verifiable Reward (RLVR), which leverages rule-based reward mechanisms to optimize models in a highly efficient and reliable manner. This approach has demonstrated remarkable success in enhancing the capabilities of large language models (LLMs) even with limited training data. Recent studies have extended this methodology to multimodal large language models (MLLMs), further showcasing its versatility. For instance, R1-V~\cite{chen2025r1v} has been applied to tasks such as geometry reasoning and visual counting, where MLLMs trained with RLVR not only exhibit strong reasoning abilities but also achieve performance comparable to Supervised Fine-Tuning (SFT) on in-domain tests, while significantly outperforming SFT models on out-of-distribution (OOD) evaluations.

In another notable work, Visual-RFT~\cite{liu2025visualrftvisualreinforcementfinetuning}, the authors validated the effectiveness of RLVR on classical computer vision tasks such as image classification and object detection. Their results demonstrated that RLVR consistently outperforms SFT across nearly all categories, highlighting its broad applicability and robustness.

Despite these advancements, the integration of RLVR with MLLMs has thus far been limited to image-text modalities. To the best of our knowledge, no prior work has explored the application of RLVR to video-based multimodal models that incorporate richer sources of information, such as audio and dynamic visual content. Bridging this gap, we present the first exploration of RLVR in conjunction with Video Omni-multimodal Models, focusing on the task of emotion recognition—a domain where both visual and audio modalities provide critical cues for accurate predictions.

In this study, we build upon HumanOmni~\cite{zhao2025humanomni}, a first open-source Omni model designed for human-centric scene understanding. By applying RLVR to HumanOmni, we aim to investigate its potential in enhancing emotion recognition performance. Our findings reveal several key insights:
\begin{itemize}
    \item Enhanced Reasoning Capability: R1-Omni demonstrate superior reasoning abilities, enabling a clearer understanding of how visual and audio information contribute to emotion recognition.
    \item Improved Understanding Capability: Compared to SFT, RLVR significantly boosts performance on emotion recognition tasks.
    \item Stronger Generalization Capability: RLVR models exhibit markedly better generalization capabilities, particularly excelling in out-of-distribution scenarios.
\end{itemize}

\section{Preliminaries}

\subsection{Reinforcement Learning with Verifiable Rewards }

\textbf{Reinforcement Learning with Verifiable Rewards} represents a novel training paradigm designed to optimize models for tasks where outcomes can be objectively verified. Examples of such tasks include mathematical problem-solving, coding challenges, and other domains with well-defined correctness criteria. Unlike traditional approaches like Reinforcement Learning from Human Feedback (RLHF), which rely on a separate reward model trained on human preferences, RLVR eliminates the need for intermediate reward modeling by directly leveraging a verification function to evaluate outputs.

At its core, RLVR simplifies the reward mechanism while ensuring alignment with the inherent correctness standards of the task. Given an input question $ q $, the policy model $ \pi_\theta $ generates a response $ o $, which is then evaluated using a verifiable reward function $ R(q, o) $. This reward function determines whether the generated output matches the ground truth, assigning a binary score:
\begin{equation}
R(q, o) =
\begin{cases} 
1, & \text{if } o = \text{ground truth}, \\
0, & \text{otherwise}.
\end{cases}
\end{equation}

The optimization objective of RLVR is formulated as follows:
\begin{equation}
\max_{\pi_\theta} \mathbb{E}_{o \sim \pi_\theta(q)} \left[ R_{\text{RLVR}}(q, o) \right],
\end{equation}
where
\begin{equation}
R_{\text{RLVR}}(q, o) = R(q, o) - \beta \cdot \text{KL}[\pi_\theta(o|q) \| \pi_{\text{ref}}(o|q)].
\end{equation}

Here, $ \pi_{\text{ref}} $ denotes the reference model prior to optimization, $ R(q, o) $ is the verifiable reward function, and $ \beta $ is a hyperparameter controlling the trade-off between maximizing the reward and maintaining proximity to the reference model via KL-divergence regularization.


In this work, we extend the application of RLVR beyond traditional domains like math and coding to the realm of multimodal emotion recognition, where both visual and audio modalities contribute to the final prediction. Specifically, we utilize the training sets from the MAFW~\cite{liu_mafw_2022} and DFEW~\cite{jiang2020dfew} datasets, comprising a total of 15,306 video samples, to train our Omni-multimodal model. Notably, these datasets only provide annotations for emotion categories, without any explicit labels or supervision for the reasoning process. Despite this limitation, by leveraging RLVR, we aim to enhance the reasoning capabilities, performance, and generalization of the Omni-multimodal model in this challenging task.

\subsection{Group Relative Policy Optimization (GRPO)}

\textbf{Group Relative Policy Optimization (GRPO)} represents a novel approach to reinforcement learning that diverges from traditional methods like Proximal Policy Optimization (PPO). Unlike PPO, which relies on a critic model to evaluate the performance of candidate policies, GRPO eliminates the need for an additional critic by directly comparing groups of generated responses. This streamlined mechanism simplifies the training process while maintaining robust optimization capabilities.

The core idea behind GRPO is to assess the relative quality of multiple candidate responses within a group. For a given input question $ q $, GRPO first generates $ G $ distinct responses $ \{o_1, o_2, \dots, o_G\} $ using the current policy $ \pi_{\theta_{\text{old}}} $. These responses are then evaluated based on their corresponding rewards $ \{r_1, r_2, \dots, r_G\} $, which are obtained through a predefined reward function. To determine the relative quality of each response, GRPO normalizes the rewards by computing their mean and standard deviation:
\begin{equation}
A_i = \frac{r_i - \text{mean}(\{r_1, \dots, r_G\})}{\text{std}(\{r_1, \dots, r_G\})},
\end{equation}
where $ A_i $ represents the normalized score indicating the relative quality of the $ i $-th response.

By leveraging this normalized scoring mechanism, GRPO encourages the model to prioritize responses with higher reward values within the group. This approach not only reduces the dependency on external critic models but also enhances the model's ability to differentiate between high-quality and low-quality outputs effectively.


 Following the approach proposed in DeepSeek R1,  we combine GRPO with RLVR to leverage the strengths of both methods. This integration allows us to achieve superior reasoning, generalization, and emotion recognition capabilities.

\section{R1-Omni}
\subsection{Cold Start with EMER Dataset}
To ensure the smooth training of RLVR, we adopt a cold start strategy inspired by the training approach used in DeepSeek R1. Specifically, we initialize the HumanOmni-0.5B by fine-tuning it on a combined dataset consisting of 232 samples from the Explainable Multimodal Emotion Reasoning (EMER)~\cite{lian2024explainablemultimodalemotionrecognition} dataset and 348 samples from our manually annotated HumanOmni dataset, resulting in a total of 580 video samples. This step is crucial for equipping the model with preliminary reasoning capabilities before proceeding to RLVR training.

The EMER dataset is specifically designed for explainable emotion reasoning tasks. It contains multimodal data (visual and audio) paired with detailed annotations that describe the reasoning process behind emotion recognition. By leveraging this dataset during the cold start phase, we enable the HumanOmni model to develop an initial understanding of how visual and audio cues contribute to emotion recognition. 
The specific format of the EMER data used during the cold start phase is presented below.
 
\begin{quote}
\textbf{Human:} \\
As an emotional recognition expert; throughout the video, which emotion conveyed by the characters is the most obvious to you? Output the thinking process in `` <think></think>'' tags and the final emotion in ``<answer></answer>'' tags.

\smallskip

\textbf{GPT:}

<think>In the video, in the opening scene, we see a male character with a slight frown, his gaze directly focused on the person opposite him, eyes wide open, seemingly speaking or expressing a certain viewpoint. In the final scene, the frown becomes more pronounced, and the male character's expression becomes more irritable. Based on these scenes, it can be inferred that the male character in this video is dissatisfied and angry with the other person. In the audio, the tone is high-pitched and emotional, with a fast pace, giving a sense of excitement. In the text, the subtitle reads: "I didn't think I was that special, I just can't understand why you're treating me like this." This sentence expresses the male character's dissatisfaction and anger. Based on the video clues of the male character's slight frown, direct gaze towards the person opposite him, wide open eyes, and the more pronounced frown and irritable expression in the final scene, it can be inferred that the male character is dissatisfied and angry with the other person. Additionally, the audio clue describing the high-pitched and emotional tone, fast pace also supports this inference. Therefore, this sentence expresses the male character's dissatisfaction and anger towards the other person.</think>

<answer>angry</answer>

\end{quote}


After completing the cold start phase, the HumanOmni model is ready for further optimization using RLVR, which will be discussed in the subsequent sections.

\subsection{RLVR}
In our framework, the RLVR training process is designed to optimize the HumanOmni-0.5B for emotion recognition tasks using multimodal inputs that include both video and audio data. 
The RLVR pipeline begins with the policy model $ \pi_\theta $, which processes the multimodal input data comprising video frames and corresponding audio streams and generates a set of candidate responses. Each response is accompanied by a detailed reasoning process, providing insights into how the model integrates visual and auditory information to arrive at its predictions. These candidate responses are then evaluated using the verifiable reward functions.

The reward function in our RLVR framework is inspired by the design principles of DeepSeek R1, which decomposes the reward into two components: \textbf{accuracy reward (\(R_{\text{acc}}\))} and \textbf{format reward (\(R_{\text{format}}\))}. These components are combined to form the total reward \(R\), defined as:
\begin{equation}
R = R_{\text{acc}} + R_{\text{format}}.
\end{equation}

\begin{figure*}
    \vspace{-5mm}
    \centering
    \includegraphics[width=1\linewidth]{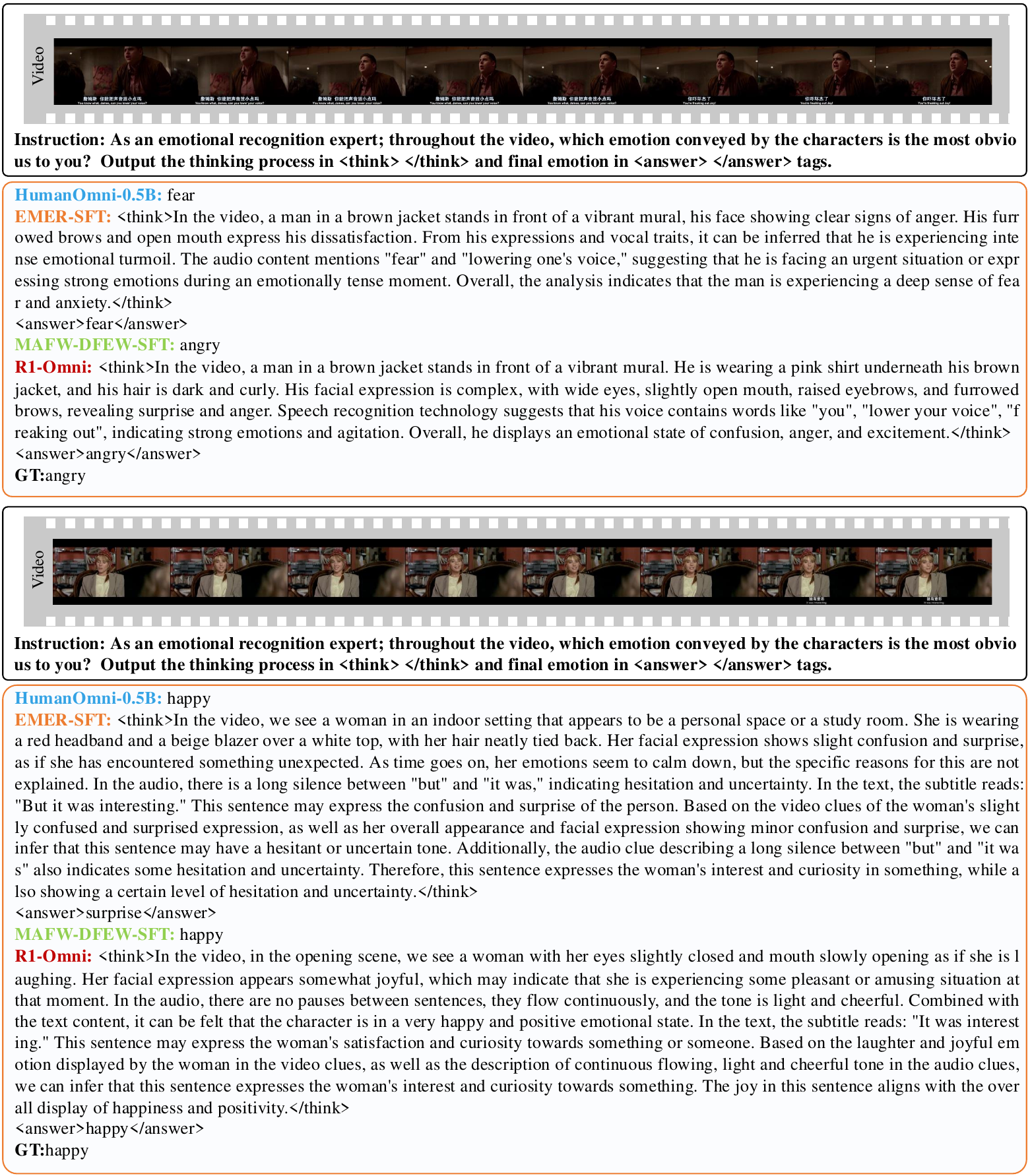}
    \vspace{-0.3cm}
    \caption{Visualization comparison.} 
    \vspace{0cm}
    \label{fig: pipeline} 
\end{figure*}

The accuracy reward (\(R_{\text{acc}}\)) evaluates the correctness of the predicted emotion compared to the ground truth (GT). To generate predictions, we use a specific prompt: "As an emotional recognition expert; throughout the video, which emotion is the most obvious to you? Output the thinking process in <think> 
</think> and final emotion in <answer> </answer> tags."

The model's output is expected to include two parts. A reasoning process enclosed within ``<think></think>'' tags, explaining how the model integrates visual and audio cues to arrive at its prediction. A final emotion label enclosed within ``<answer></answer>'' tags, representing the predicted emotion.

The accuracy reward is computed as follows:
\begin{equation}
R_{\text{acc}} =
\begin{cases} 
1, & \text{if the predicted emotion matches the ground truth}, \\
0, & \text{otherwise}.
\end{cases}
\end{equation}
This binary scoring mechanism ensures that the model is directly incentivized to produce correct emotion predictions.

The format reward enforces strict adherence to the required output structure, ensuring that the model's predictions conform to the specified HTML-like tag format. 

If the output satisfies these formatting constraints, the format reward is assigned a value of 1; otherwise, it is set to 0:

This constraint ensures that the model generates structured and interpretable outputs, facilitating downstream analysis and evaluation.

By combining these two components, the reward function not only encourages the model to produce accurate predictions but also ensures that the outputs are well-structured and aligned with the desired format.


\begin{figure*}
    \vspace{-5mm}
    \centering
    \includegraphics[width=1\linewidth]{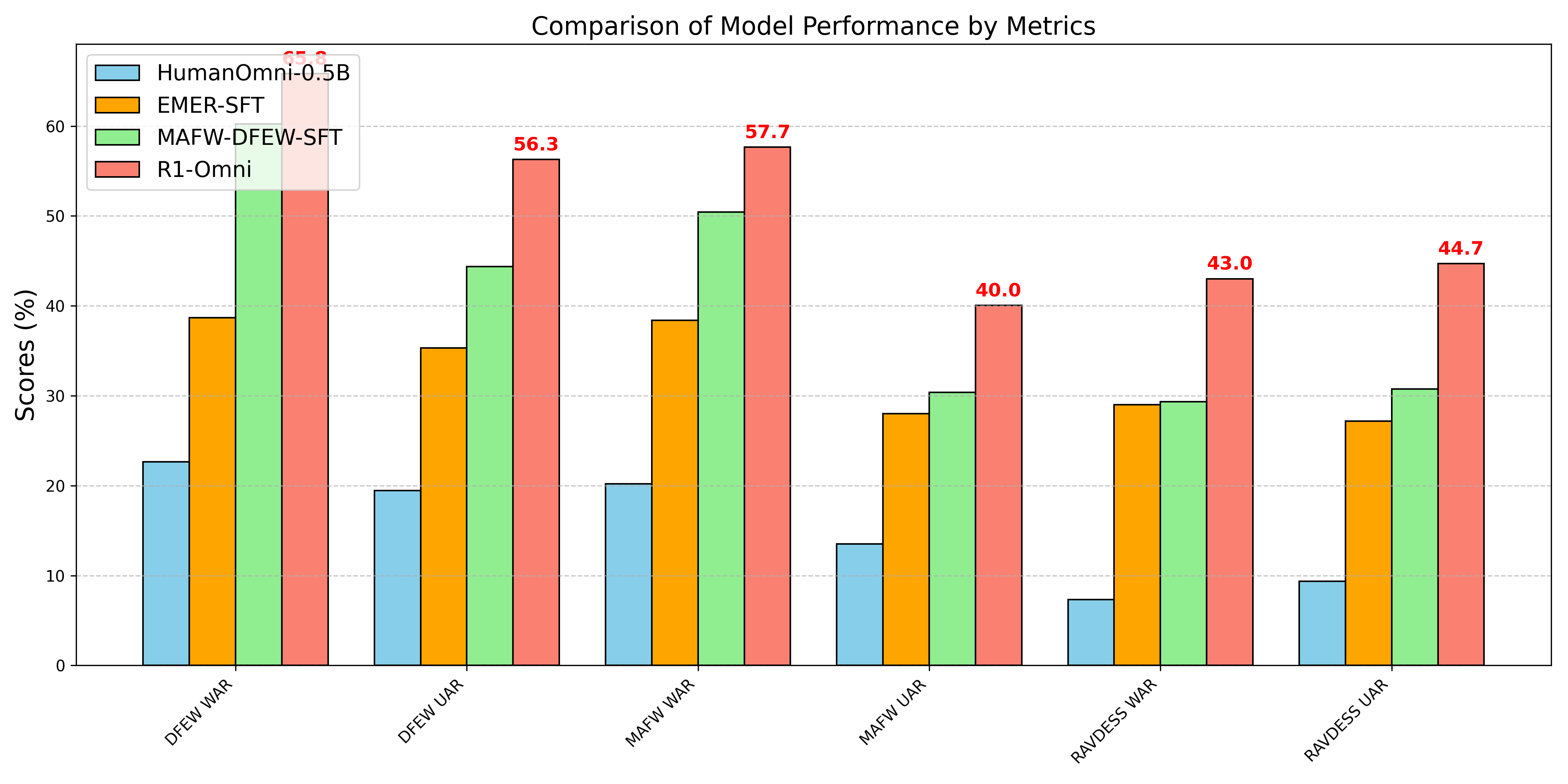}
    \vspace{-0.9cm}
    \caption{Performance comparison of models on emotion recognition datasets.} 
    \vspace{-0.2cm}
    \label{fig: comparison} 
\end{figure*}

\section{Experiments}
In this section, we present the experimental results to validate the effectiveness of our RLVR in enhancing the capabilities of the HumanOmni-0.5B. We compare R1-Omni (RLVR-trained) with three models. 1. HumanOmni-0.5B. 2.EMER-SFT: A Supervised Fine-Tuning model trained on the EMER dataset. (cold-start phase). 3. MAFW-DFEW-SFT: A Supervised Fine-Tuning model trained directly on the MAFW and DFEW training set based on HumanOmni-0.5B.

Our experiments systematically evaluate the performance of the R1-Omni and reveal three key strengths: (1) it demonstrates enhanced reasoning capability by generating detailed and interpretable explanations for its predictions; (2) it achieves improved understanding of multimodal data, resulting in higher accuracy in emotion recognition tasks; and (3) it exhibits stronger generalization to out-of-distribution data, showcasing robustness across diverse scenarios.

\subsection{Enhanced Reasoning Capability}

One of the most significant advantages of our R1-Omni is its superior reasoning ability.
To illustrate the reasoning capability of R1-Omni, we present a set of visualization examples in Figure~\ref{fig: pipeline}, comparing its outputs with those of three other models. These examples provide an intuitive sense of how R1-Omni performs relative to other approaches in terms of reasoning coherence and interpretability.

We can observe that the original HumanOmni-0.5B and the MAFW-DFEW-SFT models exhibit limited reasoning capabilities. While EMER-SFT demonstrates some level of reasoning ability, its reasoning process suffers from relatively poor coherence and is prone to generating hallucinations. For example, when tested on the MAFW and DFEW datasets, the R1-Omni  consistently outperforms these baselines by providing more coherent, accurate, and interpretable reasoning processes. This enhanced reasoning capability not only improves the model's overall performance but also offers deeper insights into how visual and audio information interact in emotion recognition tasks.

\subsection{Improved Understanding Capability}

To quantitatively evaluate the understanding capability of the R1-Omni, we compared its performance with other models on the MAFW and DFEW datasets. The metrics used for evaluation are Unweighted Average Recall (UAR) and Weighted Average Recall (WAR), which measure the model's ability to accurately classify emotions across different classes. Importantly, all evaluations were conducted using an open-vocabulary emotion testing (OV-emotion) protocol. In this setting, the model is not provided with predefined emotion categories but instead generates the emotion labels directly from the input data.

The results of the UAR and WAR metrics on the MAFW and DFEW datasets are summarized in Table~\ref{tab: human benchmark}. 

For a more intuitive comparison, please refer to Figure~\ref{fig: comparison}.

From the results presented in Table~\ref{tab: human benchmark}, we observe that The R1-Omni consistently outperforms other models on both datasets, achieving the highest UAR and WAR scores. 
The experimental results highlight the superior performance of the R1-Omni model compared to the SFT on MAFW and DFEW model. On the DFEW dataset, the R1-Omni achieves a UAR of 65.83\% and a WAR of 56.27\%, surpassing the SFT model's performance of 60.23\% UAR and 44.39\% WAR. Similarly, on the MAFW dataset, the R1-Omni demonstrates significant improvements with a UAR of 57.68\% and a WAR of 40.04\%, outperforming the SFT model, which achieves only 50.44\% UAR and 30.39\% WAR. These results clearly demonstrate that the RLVR approach not only leverages task-specific data more effectively but also enhances the model's general understanding and reasoning capabilities, leading to consistently higher performance across both datasets.

\subsection{Stronger Generalization Capability}

\begin{table*}[t]
    \centering
    \definecolor{lightlightgray}{gray}{0.8}
    \tablestyle{12pt}{1.1}
    \begin{tabular}{lcccccc}
        \toprule
        \multirow{2}{*}{\textbf{Method}}  & \multicolumn{2}{c}{\textbf{DFEW}} & \multicolumn{2}{c}{\textbf{MAFW}} & \multicolumn{2}{c}{\textbf{RAVDESS}} \\
        \cmidrule(lr){2-3} \cmidrule(lr){4-5} \cmidrule(lr){6-7} 
       & \textbf{WAR} & \textbf{UAR} & \textbf{WAR} & \textbf{UAR} & \textbf{WAR} & \textbf{UAR} \\
        \midrule

        HumanOmni-0.5B&  22.64 & 19.44 & 20.18 & 13.52& 7.33 & 9.38\\   
        EMER-SFT&  38.66 & 35.31 & 38.39 & 28.02& 29.00 & 27.19\\    
        MAFW-DFEW-SFT&  60.23 & 44.39 & 50.44 & 30.39& 29.33 & 30.75\\    
        R1-Omni&  65.83 & 56.27 & 57.68 & 40.04& 43.00& 44.69\\  
        \bottomrule
    \end{tabular}
    \caption{Results on emotion recognition datasets.}
    \label{tab: human benchmark}
\end{table*}

To evaluate the generalization capability of the R1-Omni, we conducted experiments on the RAVDESS~\cite{10.1371/journal.pone.0196391} dataset, which serves as an out-of-distribution (OOD) test set. Unlike the MAFW and DFEW datasets, which consist primarily of movie clips, the RAVDESS dataset features professional actors vocalizing lexically-matched statements in a neutral North American accent. This significant difference in data distribution makes RAVDESS an ideal benchmark for assessing the model's ability to generalize to unseen scenarios.

From the experimental results presented in Table~\ref{tab: human benchmark}, it is evident that the R1-Omni model demonstrates a substantial improvement over the MAFW-DFEW-SFT model on the RAVDESS dataset. Specifically, the R1-Omni model achieves a UAR of 43.00\% and a WAR of 44.69\%, surpassing the SFT model's performance of 29.33\% UAR and 30.75\% WAR. This significant gap highlights the superior generalization capability of the R1-Omni.



\section{Limitations}
\begin{figure*}
    \vspace{-5mm}
    \centering
    \includegraphics[width=1\linewidth]{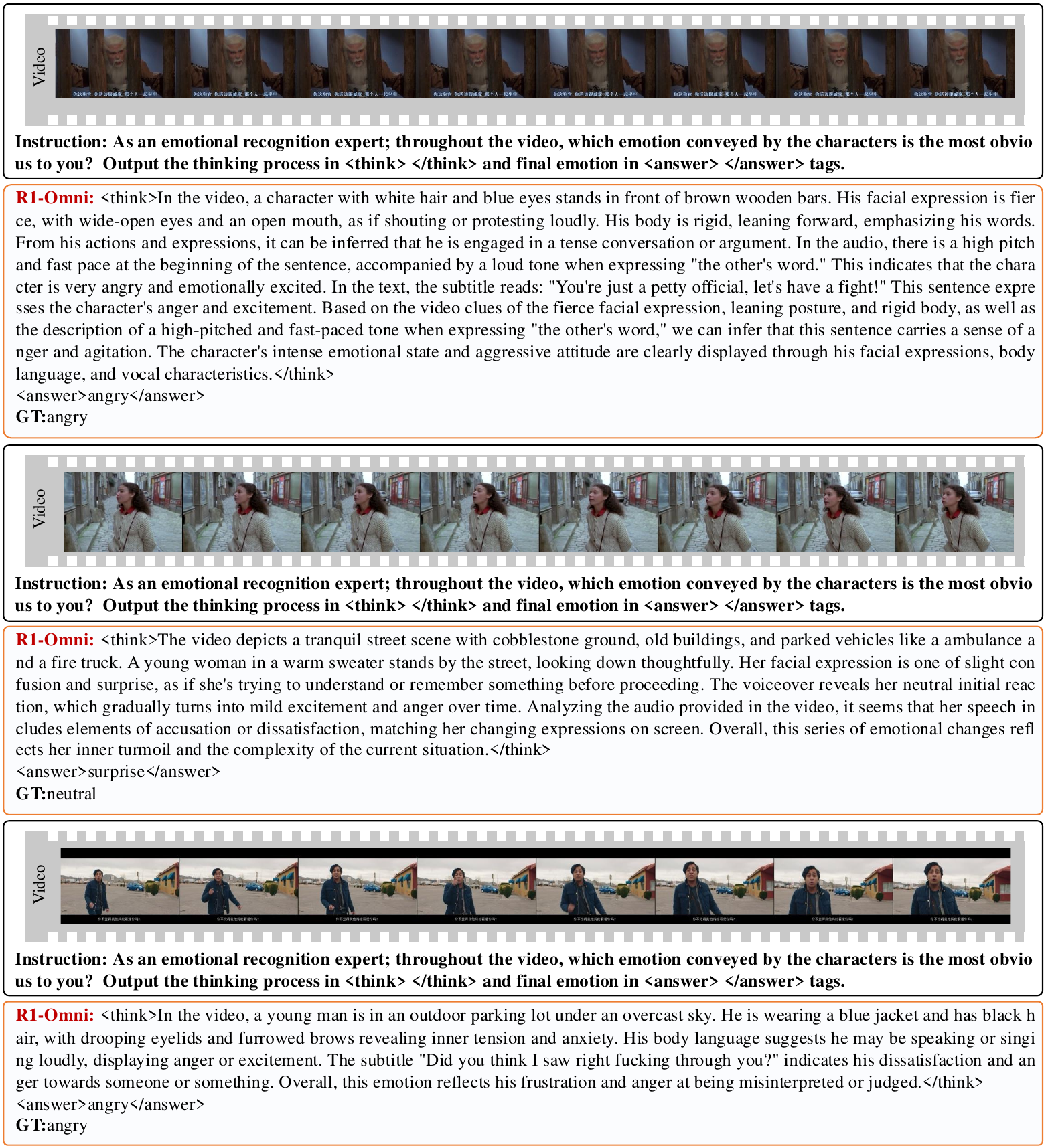}
    \vspace{-0.3cm}
    \caption{Challenging Cases.} 
    \vspace{0cm}
    \label{fig: limitation} 
\end{figure*} 

Despite the significant improvements achieved by the R1-Omni, there remain several limitations that warrant further investigation. To illustrate these challenges, we present three representative examples in Figure~\ref{fig: limitation}. 

\subsection{Inaccurate Subtitle Recognition}
In the first example, although the model produces a correct emotion prediction, we observe that inaccuracies in subtitle recognition remain a potential limitation.
This issue arises because neither the HumanOmni base model nor the subsequent SFT and RLVR training processes explicitly focus on improving subtitle recognition capabilities. Addressing this limitation will require integrating more robust subtitle processing techniques, such as fine-tuning on specialized datasets or incorporating advanced natural language understanding models.

\subsection{Hallucination in Reasoning}
The second example demonstrates a common issue hallucination, where the model generates reasoning outputs that are not grounded in the actual content of the video. For instance, the statement ``The voiceover reveals her neutral initial reaction, which gradually turns into mild excitement and anger over time'' does not align with the video's actual emotional trajectory. This fabricated reasoning leads the model to incorrectly predict the emotion as surprise, highlighting the need for mechanisms to ensure the model's outputs remain faithful to the input data. 

\subsection{Underutilization of Audio Cues}
The third example underscores the model's limited ability to fully utilize audio cues, such as tone and intonation, which are critical for accurate emotion recognition. Although our model is capable of reasoning about emotions by integrating both audio and visual information, it appears that in certain cases, the use of audio features is not as thorough or effective as the use of visual cues. In this specific instance, the character's vocal delivery provides strong emotional signals, yet the model fails to adequately incorporate these nuances into its reasoning process. 

\subsection{Implications for Future Research}

The limitations identified in our analysis highlight several promising directions for future research to further enhance the capabilities of R1-Omni. Specifically, we propose the following key areas of exploration:

\begin{enumerate}
    \item Strengthening the Foundation Model's Capabilities \\
    While RLVR significantly enhances the reasoning and generalization abilities of the base model, the inherent performance of the foundation model remains a critical determinant of overall success. Therefore, continuous efforts to improve the underlying Omni model such as through larger-scale pretraining, more diverse datasets, or advanced architectural designs are essential to unlock the full potential of RLVR-based approaches.

    \item Mitigating Hallucination in Reasoning Outputs \\
    Due to the inherent challenges of multimodal data, such as the weaker causal relationships within video and audio tokens compared to text tokens, as well as the lack of explicit supervision for reasoning content, hallucinations can occur during the model's reasoning process.
    These inaccuracies not only degrade performance but also negatively impact user experience. Developing mechanisms to detect and mitigate hallucinations will be crucial for improving the reliability and usability of the model.

    \item Enhancing Audio Cue Utilization \\
    The underutilization of audio cues, such as tone and intonation, represents a limitation in the current model. Future work should focus on improving the model's ability to extract and integrate audio features effectively. 
    
    \item Enhancing Reasoning Depth and Emotional Intelligence \\
    The current reasoning process tends to be somewhat mechanistic, focusing primarily on directly observable features such as visual cues and audio signals. However, human emotion recognition often involves deeper psychological insights, such as understanding the motivations, intentions, or internal states of individuals. By guiding the model to explore more nuanced aspects of reasoning, such as inferring psychological activities or emotional drivers, we can elevate its emotional intelligence and enhance its ability to capture complex emotional dynamics.
    This advancement would enable the model to better simulate human-like empathy and reasoning in real-world scenarios.
\end{enumerate}

\bibliography{main}
\bibliographystyle{plain}
\end{document}